# From Radiologist Report to Image Label: Assessing Latent Dirichlet Allocation in Training Neural Networks for Orthopedic Radiograph Classification

Jakub Olczak and Max Gordon

Department of Clinical Sciences, Karolinska Institute, Stockholm, Sweden

## Abstract

**Background**: Radiography (X-rays) is the dominant modality in orthopedics, and improving the interpretation of radiographs is clinically relevant. Machine learning (ML) has revolutionized data analysis and has been applied to medicine, with some success, in the form of natural language processing (NLP) and artificial neural networks (ANN). Latent Dirichlet allocation (LDA) is an NLP method that automatically categorizes documents into topics. Successfully applying ML to orthopedic radiography could enable the creation of computer-aided decision systems for use in the clinic. We studied how an automated ML pipeline could classify orthopedic trauma radiographs from radiologist reports.
**Methods**: Wrist and ankle radiographs from Danderyd Hospital in Sweden taken between 2002 and 2015, with radiologist reports. LDA was used to create image labels for radiographs from the radiologist reports. Radiographs and labels were used to train an image recognition ANN. The ANN outcomes were manually reviewed to get an accurate estimate of the method's utility and accuracy.
**Results**: Image labels generated via LDA could successfully train the ANN. The ANN reached an accuracy between 91% and 60% compared to a gold standard, depending on the label.
**Conclusions**: We found that LDA was unsuited to label orthopedic radiographs from reports with high accuracy. However, despite this, the ANN could learn to detect some features in radiographs with high accuracy. The study also illustrates how ML and ANN can be applied to medical research.

## Introduction

Radiography (colloquially called "X-ray images") is at the core of orthopedics. Treatment decisions are mainly influenced by the fracture appearance on the radiograph and less by the patient's background [1,2]. Plain radiography's ready availability, speed, low cost, and low radiation mean it remains central even though computer tomography (CT) and magnetic resonance imaging (MRI) have become more commonplace.

The human capacity for details limits the information that can be extracted from images. The simplicity needed for two humans to give the same interpretation of an image, also known as inter-observer reliability, forces further simplifications and loss of detail. For example, wrist fractures are among the most common fractures, and distal radius fractures are especially prevalent, with 31 cases per 10,000 person-years [3]. Though it is clear that if there is no displacement of the fracture, there is no need for intervention [4,5], in other situations, there is still debate about the optimal treatment [6,7]. Some measures can predict later displacement, in which case early intervention is desirable. Extracting all available information from the image, like tilt, shortening of the bone, or the level of fragment comminution [8,9], is, therefore, vital for everyday orthopedics. However, these are complex factors to measure, and inter-observer reliability becomes an issue [10,11].

A computer-aided decision-making system (CAD) could alleviate these problems. A computer program that analyses radiographic images, finds relevant pathologies, takes desired measurements, and, based on that, makes an accurate prognosis would be a valuable tool in the clinic. It could also facilitate advances in clinical treatment by allowing for new avenues of research and making radiology more available to large-scale data mining.

A fundamental part of developing such a CAD is being able to analyze and interpret radiographic images. This usually entails using extensive collections of studies with explanatory labels to train a computer model to detect image patterns [12]. Today's annotations are in plain text reports written by radiologists at hospitals and clinics worldwide. They summarize relevant findings into descriptive reports for use by other doctors. Turning written reports into labels for a CAD can be vital. Due to the large number of labeled images needed, automating this process would be crucial. A tool for addressing this problem is machine learning (ML), broadly described as computing procedures that learn patterns, rules, and models by learning from examples rather than explicitly being told what to look for [13]. ML has emerged as a widely used tool in many areas [14–22].

*Natural Language Processing and Latent Dirichlet Allocation*

Textual medical data analysis (journals, reports, archives) is a reality of everyday clinical work and research. It usually entails spending hours manually reading and classifying texts to extract interesting information. The subfield of automated language classification, spoken or written, and interpretation is called natural language processing (NLP), where "natural" implies human [23]. NLP can significantly benefit medical research [24,25].

A collection of documents (a corpus) can be anything from magazines and social media posts to scientific journals or patient records. Texts could, for example, be broad-scoped press articles from the New York Times on topics such as politics, medicine, cooking, art, or a mix of these, or a medical journal on orthopedics. Similarly, orthopedic radiologist reports could be divided into subtopics such as fractures, wrist fractures, and wrist fractures involving the styloid.

Latent Dirichlet Allocation (LDA) is a NLP algorithm that automatically tries to identify document topics [23]. It relies on the idea that topics are associated with a set of words, and every topic will have a specific combination of words that appear more commonly than other topics. LDA assumes topics can be identified by their unique combinations of words without considering word order, [26] also called a "bag-of-words" model. [23] LDA first appeared in Blei et al. 2003 [27] and is often implemented based on Griffith and Steyvers 2004 [28]. Our usage of LDA on radiologist reports to automatically generate topics and labels was inspired by Shin et al. [29]. They used a mix of radiographs, CT, and MRI image slices from labeled via radiologist reports using LDA to obtain general results. The results were interesting and broadly illustrative, detecting a wide range of pathologies but lacked accuracy or clinical relevance. They also used online databases of medical language [30,31] to interpret topics and assign labels. Such databases of medical terms and their relationships are not publicly available in Swedish. To our knowledge, no studies have attempted large-scale machine learning for orthopedic trauma radiographs and radiologist reports in Swedish.

LDA assumes a hidden (latent) structure to the documents of the corpus and the topics and is founded on a specific idea of how documents are created. The fundamental building blocks of documents are words. LDA assumes a document is about one or more topics, e.g., a central topic and several subtopics. The words in a document make up the vocabulary for the document. Topics are defined by their specific vocabulary (words and the combinations of words) but not their order (syntax). The idea behind LDA is to reverse the document generation process to find topics. Rather than combining words from a topic into a document, it divides documents into words and examines their combination (the vocabulary) to associate it with topics. Documents with a similar vocabulary are assumed to be about the same topic. LDA tries to find these combinations of words using statistical simulations. It randomly distributes the available words over topics and tests how well the resulting word combinations fit the data. It repeatedly repeats this process, encouraging models that best fit the previous step. Over time, topics that fit the data well evolve. [26]

More theoretically, a generative model is a statistical model that generates observable data values from observations. LDA is a generative model that explains observations (vocabularies) as latent (hidden) structures and topics (the observable data values). The Dirichlet distribution is a standard probability distribution for modeling belief (what words we believe the topics are made up of) in Bayesian statistics. LDA postulates that the characteristics of the documents are Dirichlet distributed and uses the distribution to draw random words from the corpus and assigns them to topics and topics to documents. LDA tries to solve the joint probability distribution described by equation [1],

$$p(\alpha, \beta) = p(\alpha)p(\theta)p(\beta)p(z, \phi) \quad [1]$$

The parts of the equation [1] are as follows
$p(\alpha, \beta)$: the probability model when combining $\alpha$ and $\beta$ (sampling parameters for the Dirichlet distribution). $\alpha$ and $\beta$ are sampling parameters that determine how words and topics are drawn randomly during the modeling process that generates the topics.
$\alpha$: estimates to what extent documents are made up of more (large α) or fewer (small α) topics. It determines how topics are drawn and distributed over documents.
$\beta$: estimates to what extent words belong to many different topics (mixed between topics) or fewer topics. It determines how words are drawn and distributed over topics.
$\theta$: topics and how they are distributed over the documents.
$\phi$: words and how they are distributed over topics.
$z$: words randomly drawn from the probability distributions of $\phi$.

*Artificial Neural Networks for Image Analysis*

Artificial neural networks (ANN) have become widespread for image analysis. It has revolutionized the field of image recognition and enabled computers to perform at superhuman capacity for specific image recognition tasks [32]. Automatic detection and classification of pathologies in radiographs and CAD have a wide range of potential uses. Automatic detection could be a screening tool in the emergency room or facilitate radiologist work. It could also be helpful when radiologists are rare, for example, in triage during natural disasters or in parts of the world where doctors are generally scarce. If the method works, it could be extended to other fields, such as fracture outcome prediction or mammography analysis.

We aimed to study how an automated ML pipeline could classify orthopedic trauma radiographs from radiologist reports. The hypotheses were that 1) it is possible to automatically find clinically relevant image features in radiologist reports using NLP, and 2) these labels could be used to train an image recognition ANN. The detection accuracy of the ANN, trained with the labels created using NLP, was the primary outcome.

## Methods and Materials

This was a cross-sectional study of patients examined at Danderyd Hospital, Stockholm, Sweden, from 2002 to 2015. All patients with a radiologist report about wrist or ankle radiography examinations available in the Radiology Information System (RIS) were eligible for inclusion. Examinations not containing radiographic images and radiologist reports were excluded, as were examinations where reports were less than six characters, as they could not contain any diagnostic information. Table 1 displays statistics for the data included in the study.

**Table 1: Image and report data used in the experiments.** Experiment 1 used randomly drawn reports from the wrist reports. And did not use any radiographs. Experiment 2 used reports to construct image labels for radiographs and train an ANN to detect them. Experiment 3 used random radiographs from Experiment 2 with their ANN outcomes.

|  | **Experiment 1** | **Experiment 2** | | **Experiment 3** |
|---|---|---|---|---|
| **Body part** | Wrist | Ankle | Wrist | Ankle & wrist |
| **Documents** | 24,948 random reports | 88,026 labeled reports | | – |
| **Documents used** | 44,890 sentences and 24,948 reports | 76,426 sentences | 102,930 sentences | – |
| **Radiographs** | – | 234,870 | | 1,500 (300 random images per label, for 5 labels) |

The reports were cleaned of identifiable information. Any names, locations, radiologist information, and Swedish personal identity numbers were removed. Reference IDs for other examinations and dates were replaced by placeholder text stating that either the examination ID or date had been removed. Text not about wrist or ankle radiographs was removed from the reports. The corpus documents were corrected for spelling and regularized; for example, many synonyms were replaced with one word to make it more homogenous and less variable using a manually created list of ~10,000 corrections specially developed for the corpus. We then applied word stemming (reducing words to the basic word stems); very rare words, nonsensical text strings, and 112 stop words were removed from the corpus (using the SnowballC package in R for the Swedish language, but not the words "no" and "none" ("inget" and "ingen"). Stop words are words that are so very frequent in natural language that they would block out other signals or are largely irrelevant, like "me," "you," "his," "hers," "to," etcetera. Lastly, we removed non-word characters (all except letters, blank space, scores [—], and underscore [ _ ]), divided the documents into words (tokenizing), and then removed short documents that could have arisen from our text preprocessing.

## Experiment Design

The study consisted of three experiments: 1) calibrate LDA to our data, 3) create LDA image labels and implement the ANN, and 3) evaluate the true accuracy of the ANN and the feasibility of using LDA labels by comparing ANN outcomes to a gold standard (Table 2, and Figure 1.) This study used an image recognition ANN in the form of a convolutional neural network (CNN).

**Table 2: Summary of experiments.**

*Experiment 1 summary*

> **Experiment aim:** Calibrate LDA topic modeling for radiologist report text.
>
> **Study subjects:** 24,948 randomly drawn radiologist reports of wrist radiograph examinations.
>
> **Exposure variables:** $\alpha$ (four values via a scaling factor), document type (report or sentence), presentation (top words, top documents or both).
>
> **Outcome:** A linear regression model describing how to best model radiologist reports for our study.

*Experiment 2 summary*

> **Experiment aim:** Use LDA to generate clinically relevant labels that describe the contents of the radiologist report and use those that describe image features to label radiograph images. Use labels to train an image recognition ANN to detect the label features in radiographs.
>
> **Study subjects:** 179,356 sentences from 88,026 radiologist reports on wrist and ankle radiograph examinations, with 234,870 radiographs.
>
> **Exposure variables:** Sentences were exposed to LDA, with parameters selected in experiment 1, to generate descriptive labels for associated radiographs. Labeled radiographs were used to train an image recognition ANN. The trained ANN was exposed to a test set of labeled radiographs.
>
> **Outcome:** Accuracy of the ANN image recognition outcome compared to LDA outcome for each label, and five labels selected for further study.

*Experiment 3 summary*

> **Experiment aim:** Evaluate the quality of the ANN image recognition labeling outcomes, from experiment 2, by comparing them to a gold standard.
>
> **Study subjects:** 1,500 radiographic images: 300 labeled image (150 misclassified and 150 correctly classified) per label, for the five ANN labels from experiment 2.
>
> **Exposure variables:** Five ANN label outcomes from experiment 2 compared to a gold standard.
>
> **Outcome:** Revised accuracy as an estimate of the true accuracy of the ANN.

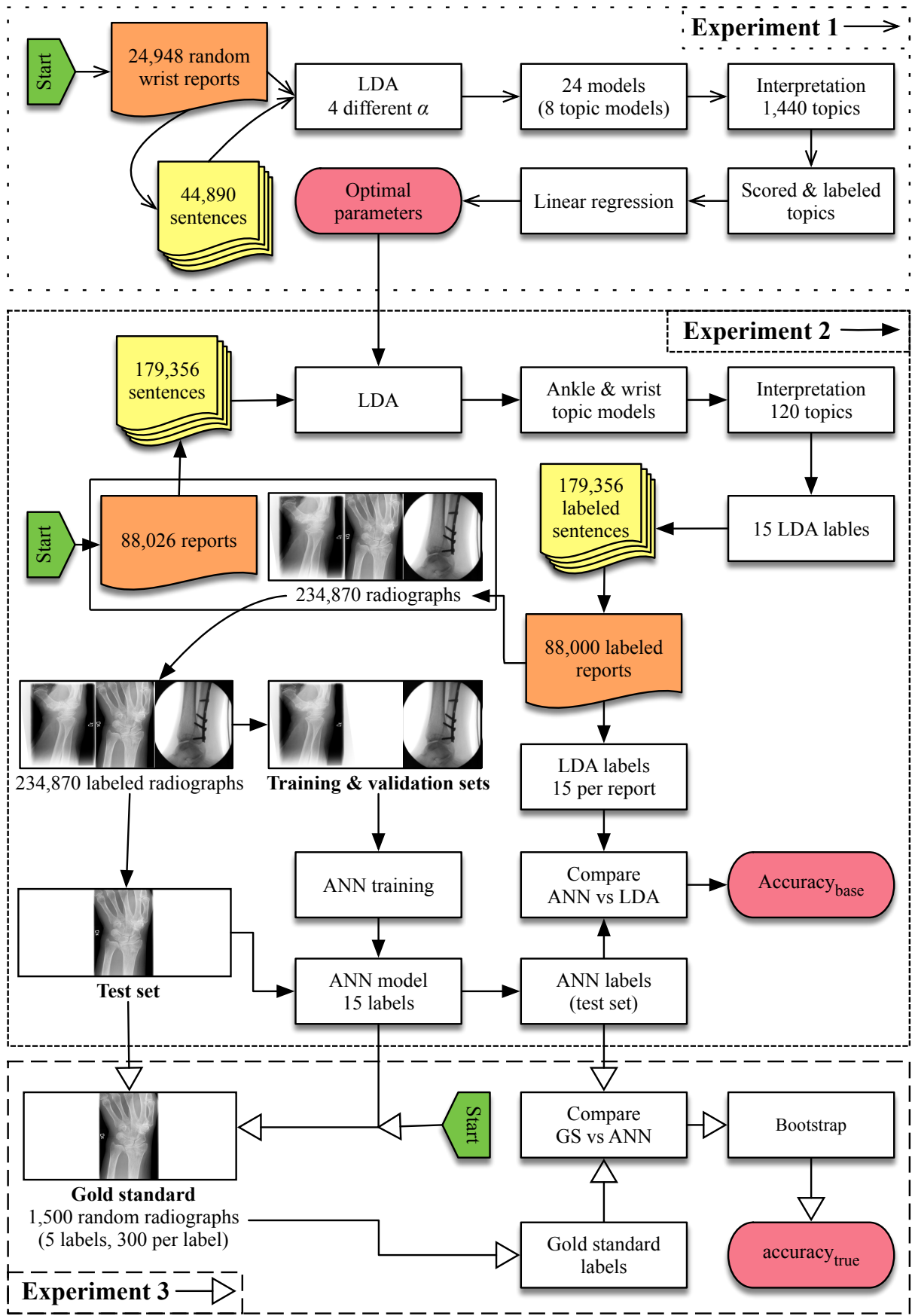

**Figure 1. Experiment flowchart with interdependencies.** Experiment startingpoints are colored green, reports are yellow, sentences orange and desired outcomes are red endpoints.
Abbreviations: ANN: Artificial neural network. GS: gold standard. LDA: Latent Dirichlet Allocation.

*Experiment 1 – Topic Modeling Parameter Study*

**Aim**: Study how to select modeling parameters to get a good topic model for the radiologist reports.

The LDA parameter α estimates how many topics a document is assumed to deal with and will vary depending on the corpus. α is usually set to $\alpha = \frac{50}{number\ of\ topics}$, where the number of topics is either a convenient number or estimated by some mathematical means. It will depend on the corpus. This selection of α is only based on an assumed variability in topics and does not consider the nature or number of the documents being analyzed. We stipulated that the nature of the documents matters. Our corpus was a collection of radiologist reports. The radiologist reports were generally concise and consisted of short, independent sentences describing a distinct image feature (e.g., fracture, dislocation, arthrosis, etc.) The vocabulary should be limited and highly domain-dependent, as should the number of topics discussed for a specific type of examination. The number of topics per report should also be limited as they only discussed findings in the image.

**Study subjects:** We chose 25,000 random wrist examination radiologist reports (Table 1), which were further divided into individual sentences.

We redefined α as in equation [2],

$$\alpha = scaling\ factor \cdot \frac{50}{number\ of\ topics}, \qquad [2]$$

where the scaling factor is just a multiplier that allows α to vary without varying the number of topics, i.e. change how specifically on-topic documents were without adding more topics. Computing equation [1] required the computation of the expression $p(w|\alpha, \beta)$, where $w$ represents the words in the vocabulary, and the literature has different suggestions for doing this. We used collapsed Gibbs sampling, as in Griffiths and Styvers 2004 [28]. The resulting LDA model consisted of $\theta$ and $\phi$, the topic distribution for each document and word distribution for each topic, respectively, where the "distribution" is a probability for each document or word belonging to a particular topic. I.e., for each document, we can get the probability that it belongs to a specific topic from $\theta$, and for each word, we can get the probability of it belonging to a particular topic from $\phi$.

**Exposure:** We chose four scaling factors and two different corpora (reports and sentences), resulting in eight different LDA models. We tabulated all models in three different ways: 1) - top words (highest probability) per topic with a minimum 3% incidence in the topic, 2) - top 15 documents (reports or sentences) per topic, and 3) - both together. This effectively turned one LDA model into three models, presenting different information on which to make conclusions. As each LDA model was presented in three different ways, we got 24 models.

Topics were randomly reordered, and all models were given, blinded, to a resident radiologist (fourth-year resident). The radiologist was tasked with determining a description for each topic based on the information in each model and assigning a score of 0 or 1 to 10 for the quality of the description. 0 (zero) meant that it was not possible to give an interpretation of the topic. A score of 1 was low, which meant the radiologist could barely guess what the topic could be, and a score of 10 meant the description was clear and deemed to fit the data flawlessly.

Using that data, we performed a linear regression. From the linear regression, we selected the best topic model, which had the most useful and interpretable image labels.

**Outcome:** For each model, a median, mean score, standard deviation, standard error of the mean, and the number of unique topics were computed (some topics reoccurred). The models were ranked by mean score, unique topics, and median score in order of decreasing importance. Linear regression modeling was used to determine how the mean score depended on scaling factor, document type (report or sentence), view (presented as top documents, top words, or both), and number of unique topics.

*Experiment 2 – Image Labeling and Image Recognition*

**Aim:** Generate image labels describing the contents of the radiographs. These labels were meant to be features in the radiographs, i.e., what the radiologist had seen and described in the report, and they needed to be detectable in the radiograph by the CNN.

**Study subjects:** We used reports from radiographic examinations of wrists and ankles taken at Danderyd Hospital between 2002 and 2015. Data are displayed in Table 1.

Utilizing the best model parameters from experiment 1, we generated one topic model per anatomy (wrist and ankle) with 100 topics each. The topics generated by LDA were analyzed, and topic descriptions were given. For every document, every topic was either true – if it had sufficiently high probability ($p > 1/20$) for that document – or no answer (missing) if not ($\theta$ in equation [1]). The cutoff was selected as it meant it was five times more likely than pure chance (1/100) to belong to that topic. Image labels, i.e., properties/features of the image, were deduced from the topic descriptions. By merging related topics and labels, a false result was introduced. For example, "fracture" and "no fracture" are mutually exclusive. If the "fracture" label was true, then the "no fracture" label could not be true for the same document and vice versa. Combining labels into a single label gave three possible outcomes: true, false, and missing. Labels for each anatomy were merged into labels for the data. Whenever there was no information, the outcome was set to missing. This was done to not skew the ratio between true and false outcomes to a degree where constantly guessing false or true gives a high accuracy simply by being much more common. If a value is false 95% of the time, continually guessing false would result in 95% accuracy. Missing values were excluded from the analysis. The most common outcome and frequency of the most common outcome was called the best guess (or the mode in statistical literature), and we derived the mode for each label.

**Exposure variable:** The report labels were used as labels for all radiographs associated with the examination, where each examination generally contained two or more images. Images were randomly divided into three groups. 70% of the images were reserved for training the CNN to detect image features. 20% of the images were used for verification, i.e., to give a quality estimate for the model during training. 10% was set aside for testing the trained CNN model, i.e., computing the accuracy of the CNN for each label. The four possible outcomes are described in Table 3.

**Outcome:** Accuracy meant the percentage of cases where the CNN image label corresponded to the LDA label (TN+TP). We called this the base accuracy, $accuracy_{base}$ of the CNN, and it was computed as in equation [3],

$$accuracy_{base} = \frac{True\ positives + True\ negatives}{number\ of\ images\ in\ test\ set} = \frac{TP+TN}{TN+FN+TP+TN}. \quad [3]$$

We also compared the $accuracy_{base}$ to the mode to get a sense of the label quality. If $accuracy_{base}$ is smaller than the mode, we would get more correct answers using the mode instead of our model.

**Table 3: Confusion matrix of LDA vs CNN outcomes.** Classification outcomes for labels when testing the CNN. The feature described by the label was either present (TRUE) or not present (FALSE) in the document/image. A third outcome was missing when there was no information about the label for the image/report. The missing outcomes were discarded from CNN training and analysis.

|  |  | CNN outcome | |
|---|---|---|---|
|  |  | **TRUE** | **FALSE** |
| **LDA outcome** | **TRUE** | True positive (TP) | False positive (FP) |
|  | **FALSE** | False negative (TN) | True negative (TN) |

*Experiment 3 - Reviewing neural network classifications*

**Aim:** Evaluate the quality and usefulness of the image labels in experiment 2, where the base accuracy was calculated by comparing the LDA label to the CNN label (as in Table 3 and equation [3]).

**Study subjects:** Five labels generated by LDA in experiment 1 were selected for study in experiment 3. 234 images were needed for a power of 0.8 with p=0.95 to compare CNN to the standard treatment (manual review). We studied 300 images per label.

**Exposure variables:** The LDA label value was derived from text, and the CNN label values were derived from examining images, so the base accuracy compares two different kinds of outcomes (document-derived versus image-derived). Therefore, it made sense to study how well these agree.

The base accuracy, $accuracy_{base}$, estimates the CNN's actual accuracy or performance in detecting a particular image feature. If the label that LDA generates is 1) incorrect (for example, "no fracture" is labeled as a "fracture"), 2) incomplete (arthritis is not mentioned in the report but is present in the image), or 3) partially correct (the label is visible in one but not all images in the examination) the CNN could correctly fail/succeed to detect this in the image. The CNN classification would be correct, but it will come up with an error (FN or FP) because it is inconsistent with the LDA-generated image label. In case 3) both LDA and CNN are correct but could not be correct for all images in the examination. We wanted to study these errors to assess the method's effectiveness correctly. The error (per label), base error, or $error_{base}$ was $error_{base} = 1 - accuracy_{base}$.

The base error was the proportion of images where the CNN classification differed from the LDA label (misclassified or misses), $proportion_{misses}$. The base accuracy was the proportion of images where the CNN agreed with the label (hits) $proportion_{hits}$.

To get an accurate estimate of the CNN error, we randomly drew 150 correctly classified radiographs and 150 misclassified for each label. The images were manually reviewed to see if the image label was present or not in the image, giving a gold standard. The CNN outcomes were compared to the gold standard giving an accuracy for the hits and misses, $accuracy_{hits}$ and $accuracy_{miss}$ respectively. An estimate of the CNN's true accuracy, $accuracy_{true}$, was given by computing a weighted accuracy as in equation [4],

$$accuracy_{true} = accuracy_{miss} \cdot proportion_{miss} + accuracy_{hit} \cdot proportion_{hit}. \quad [4]$$

We used non-parametric bootstrapping to compute an estimate for $accuracy_{true}$ a 95% confidence interval (95% CI) [33,34].

**Outcome:** Using bootstrapping, resampling 10,000 times each from $accuracy_{miss}$ and $accuracy_{hits}$ distributions, we got 10,000 $accuracy_{true}$ estimates from equation [4]. The $accuracy_{true}$ was given by the 50% quantile (median) and a 95%CI from the upper 2.5% and lower 97.5% percentile bounds, i.e. 2.5% percentile < median < 97.5% percentile.

## Technological Considerations

We used freely available open-source implementations of algorithms and software. LDA was implemented via the topicmodel package [35], version 0.2-4, for the R statistical programming language, version 3.3.0. The VGG-16 layers CNN was implemented in the Torch7 framework [36,37]. VGG-16 was chosen because it was freely available and widely used [29,38,39]. A study comparing different CNNs showed that it performed well for our data. See Olczak et al. 2017 for further details on our implementation of VGG-16 [40].

# Results

## Experiment 1

Experiment 1 used 24,948 randomly drawn wrist reports from the dataset, resulting in 44,890 sentences. Eight different topic models were generated using four different scaling factors (0.01, 0.1, 1, and 10 in equation [2]), two different corpus (reports and sentences), and 60 topics each. Each of the eight models was combined with the top 15 documents that best fit the topic model (combination of words), all words with more than 3% incidence in the topic, or both (top words and best fitting documents). As the results of each parameter combination, scaling factor, and corpus were presented in three different ways, it resulted in 24 different models and amounted to 1440 individually labeled and scored topics. The 24 model parameters and detailed results are in Table 4.

The "normal" scaling factor gave many interpretable topics (7) but not a high mean score (5.9), even for the best model. Among our choices, the most optimal value for the scaling factor was the "small" scaling factor (0.1). It was less clear if it was best to see documents, words, or both, but the last option tended to rank higher. The best topic model was generated using individual sentences based on linear regression (Table 5), consistent with Table 4. A small scaling factor (0.1) for our corpus of medical radiologist reports gave the most interpretable model regarding the highest mean score and most unique topics. To determine topics, we should also study both reports and top words.

**Table 4: Comparison of 24 topic models.** Topics are collections of words with estimates of the probability of finding each word in the topic. The table shows how the mean score and number of unique topics found by the reviewer depended on the scaling factor multiplier, type of corpus (report or sentences), and the information the reviewer had available (words, documents, or both). Models with the same scaling factor and corpus are the same model and differ in view. Each model had 60 topics.

| Scaling factor (value) | Document Collection | View | Median | Mean | Standard deviation of mean | Standard error of mean | Unique Topic Labels |
|---|---|---|---|---|---|---|---|
| Small (0.1) | sentences | both | 8.0 | 7.4 | 3.2 | 0.4 | 38 |
| Small (0.1) | sentences | docs | 8.0 | 6.5 | 3.4 | 0.4 | 31 |
| Tiny (0.01) | sentences | both | 8.0 | 6.4 | 3.7 | 0.5 | 33 |
| Tiny (0.01) | report | both | 8.0 | 6.4 | 3.7 | 0.5 | 32 |
| Small (0.1) | report | both | 8.0 | 6.3 | 3.6 | 0.5 | 29 |
| Small (0.1) | report | words | 7.0 | 6.2 | 3.4 | 0.4 | 27 |
| Tiny (0.01) | sentences | words | 7.0 | 6.0 | 3.0 | 0.4 | 31 |
| Normal (1) | sentences | both | 7.0 | 5.9 | 3.4 | 0.4 | 33 |
| Normal (1) | sentences | docs | 6.0 | 5.7 | 3.5 | 0.5 | 30 |
| Tiny (0.01) | sentences | docs | 7.5 | 5.7 | 3.8 | 0.5 | 26 |
| Normal (1) | report | words | 5.0 | 5.6 | 3.6 | 0.5 | 40 |
| Large (10) | report | words | 6.0 | 5.1 | 3.7 | 0.5 | 26 |
| Small (0.1) | report | docs | 5.0 | 5.0 | 3.6 | 0.5 | 28 |
| Tiny (0.1) | report | words | 5.0 | 4.6 | 2.9 | 0.4 | 37 |
| Tiny (0.1) | report | docs | 4.0 | 4.4 | 3.5 | 0.5 | 30 |
| Small (0.1) | sentences | words | 5.0 | 4.3 | 3.5 | 0.4 | 24 |
| Normal (1) | sentences | words | 3.0 | 4.3 | 4.0 | 0.5 | 22 |
| Large (10) | sentences | both | 3.0 | 4.2 | 4.0 | 0.5 | 20 |
| Large (10) | sentences | words | 3.0 | 3.8 | 3.9 | 0.5 | 22 |
| Normal (1) | report | both | 2.0 | 3.5 | 3.4 | 0.4 | 28 |
| Large (10) | report | both | 2.0 | 3.2 | 3.2 | 0.4 | 22 |
| Large (10) | sentences | docs | 0.0 | 3.0 | 3.9 | 0.5 | 18 |
| Normal (1) | report | docs | 1.0 | 2.3 | 3.2 | 0.4 | 20 |
| Large (10) | report | docs | 0.5 | 1.7 | 2.6 | 0.3 | 14 |

**Table 5: Linear regression model of mean score.** "Crude" is the univariate analysis model, and "adjusted" is the multivariate analysis, which uses all variables in the table.

|  | Crude | | Adjusted | |
|---|---|---|---|---|
| **Variable** | **Coefficient** | **2.5% to 97.5%** | **Coefficient** | **2.5% to 97.5%** |
| **Intercept** | 4.90 | 4.28 to 5.51 | 0.40 | -1.62 to 2.41 |
| **Unique labels** | 0.17 | 0.11 to 0.24 | 0.15 | 0.06 to 0.23 |
| **Scaling Factor (value) for α in equation [2]** | | | | |
| **Large (10)** | 0.00 | Reference | 0.00 | Reference |
| **Normal (1)** | 1.05 | -0.36 to 2.46 | -0.18 | -1.36 to 1.00 |
| **Small (.1)** | 2.45 | 1.04 to 3.86 | 1.12 | -0.09 to 2.33 |
| **Tiny (.01)** | 2.08 | 0.68 to 3.49 | 0.46 | -0.85 to 1.78 |
| **Document type for LDA modeling** | | | | |
| **Report** | 0.00 | Reference | 0.00 | Reference |
| **Sentences** | 0.74 | -0.48 to 1.96 | 0.80 | 0.12 to 1.48 |
| **Information visible to topic reviewer** | | | | |
| **Both** | 0.00 | Reference | 0.00 | Reference |
| **Documents** | -1.12 | -2.63 to 0.38 | -0.44 | -1.35 to 0.48 |
| **Words** | -0.43 | -1.93 to 1.08 | -0.32 | -1.15 to 0.52 |

## Experiment 2

Using the results from experiment 1, the corpus was divided into sentences, and we used a small scaling factor (equal to 0.1 in equation [2]) to generate topic models, one for each anatomy separately. Labeling topics was done using a combination of top words and top documents, as was suggested by experiment 1. Similar topics were merged into a single label, and we extracted 15 labels of clinical interest. As some labels were never deduced for an examination type (for example, "fracture of the lateral malleolus" was not deduced for wrist reports), these received the value missing in the merged data set.

Results for each sentence were aggregated into results for the report by assigning the same value for each label in the report as its sentences had. If there was a conflict and the sentence values did not agree, a true value (the image feature was mentioned in a sentence) took precedence over false (the feature was explicitly stated as not present in a sentence) and missing (not mentioned in a sentence), and false took precedence over missing. For example, the first sentence of a report might say, "Intraarticular fracture of the distal radius." (Fracture and radius fracture = true), and the next, "No additional fractures" (Fracture = false). This would result in fracture = true and radius fracture = true for the report and image. As "osteosynthesis device" (OD) was not mentioned, that label was missing.

The original 179,356 sentences resulted in 15 image labels for 88,026 labeled reports (each describing a single examination) with 234,870 labeled images (each examination generally having multiple radiographs). Labels with statistics, except "no fracture", are found in Table 6. Five representative LDA labels were selected for further study, and their base accuracy is found under Experiment 2 in Table 7.

The labels were selected to be illustrative and representative and to encompass a high and low base accuracy. They were also deemed to be easily verifiable in a radiograph. Fracture and radius fractures were selected because CNN base accuracy was considerably better than the mode (17% and 15%, respectively). OD was chosen because it should be easy to spot in radiographs and performed close to the mode. Like fractures, OD was a label that would be found in both ankle and wrist images.

**Table 6: Labels with statistics.** Topic labels with LDA and CNN accuracy. The number of documents and images labeled. These are tabulated along with the most common outcome (mode) for each image label, the frequency of the mode, and CNN accuracy. No. pos.: number of reports or images labeled as true for the label. No. neg.: number of reports or images labeled as false for the label.

| Image label | No. pos. documents | No. neg. documents | No. pos. images | No. neg. images | Mode | Mode frequency | CNN base accuracy |
|---|---|---|---|---|---|---|---|
| Degenerative changes | 11,946 | 27,968 | 28,273 | 94,462 | No | 77% | 79.0% |
| Fracture | 49,297 | 31,850 | 128,098 | 92,305 | Yes | 58% | 75.0% |
| Fracture, fragmented | 17,057 | 43,073 | 47,045 | 123,903 | No | 72% | 76.8% |
| Fracture dislocated | 22,953 | 50,792 | 55,031 | 145,966 | No | 73% | 81.5% |
| Fracture, fibula | 5,837 | 24,073 | 19,808 | 85,465 | No | 81% | 80.0% |
| Fracture, tibia | 3,690 | 25,162 | 89,563 | 11,908 | No | 88% | 80.1% |
| Fracture, lateral malleolus | 4,461 | 24,576 | 15,418 | 87,064 | No | 85% | 77.0% |
| Fracture, medial malleolus | 2,413 | 25,758 | 8,303 | 91,367 | No | 92% | 80.8% |
| Fracture, incongruent | 2,279 | 28,198 | 7,611 | 98,896 | No | 93% | 84.5% |
| Osteosynthesis device | 6,678 | 23,981 | 18,850 | 86,521 | No | 82% | 83.1% |
| Fracture, styloid | 6,248 | 21,963 | 47,146 | 13,832 | No | 77% | 59.0% |
| Fracture, radius | 15,192 | 21,193 | 33,165 | 45,713 | No | 58% | 73.9% |
| Fracture, radius, not dislocated | 2,865 | 39,187 | 6,065 | 84,871 | No | 93% | 89.9% |
| Fracture compressed | 12,366 | 21,833 | 26,858 | 47,243 | No | 64% | 69.4% |

The fibula fracture label was an ankle radiograph label close to the mode. The styloid fracture label was selected because it had the worst accuracy of all labels and performed much worse than the mode. Two selected outcomes, fibular and ulnar styloid fractures, had a base accuracy worse than the mode. Ulnar styloid fractures were the worst and performed slightly better than a coin flip (50/50 chance) at assessing if the sought information was present in the image. Simply guessing false would have given considerably higher accuracy. Fracture and radius fractures performed very well compared to the mode, whereas the OD class was slightly better.

**Table 7: CNN accuracy for five labels studied in experiments 2 and 3.** The mode, the frequency of the most common outcome, was compared to the $accuracy_{base}$ of the CNN. The $accuracy_{true}$ from experiment 3 is displayed with a 95%CI.

|  | Experiment 2 | | Experiment 3 | |
| --- | --- | --- | --- | --- |
| **Image label** | **Mode (%)** | **Accuracy$_{base}$** | **Accuracy$_{true}$** | **2.5% to 97.5%** |
| **Fracture** | Yes (58%) | 75.0% | 91.0% | 87.8 to 93.7 |
| **Fracture, fibula** | No (81%) | 80.0% | 88.5% | 85.2 to 91.3 |
| **Osteosynthesis device** | No (82%) | 83.1% | 85.6% | 82.2 to 88.5 |
| **Fracture, radius** | No (58%) | 73.9% | 80.3% | 76.8 to 83.8 |
| **Fracture, ulnar styloid** | No (77%) | 59.0% | 59.9% | 55.5 to 65.8 |

## Experiment 3

In experiment 2, five representative labels were selected. For each of the five labels, 150 misclassified and 150 correctly classified images were randomly drawn from the test set. These were manually reviewed to create a gold standard, and the CNN outcome was computed compared to the gold standard and $accuracy_{true}$ was computed as in equation [4] with bootstrapping. 1,500 radiographs were reviewed for the five labels, with a gold standard of 300 images per label, and CNN outcomes were compared to the gold standard. Results are presented under Experiment 3 in Table 7.

The best-performing category was fracture, *compared to the mode*, followed by radius fracture. While other categories, such as fibular fracture and OD, achieved slightly higher $accuracy_{true}$, fracture, and radius fracture had a much lower starting mode, i.e., more was learned. OD was significantly better than the mode, but just barely. Fibular fracture base performance was slightly below the mode but reached an accuracy of 88.5% after manual review. Fractures of the ulnar styloid process did not improve their accuracy with any significance.

## Discussion

This study examined how automated ML could classify orthopedic trauma radiographs from radiologist reports and then use those labels to train an image classification CNN. We found that 1) it was possible to automatically find clinically relevant information in radiologist reports using NLP, although our data did not conform well to the standard premises of LDA. 2) The LDA labels could be used to train a CNN to detect this information in the radiographs. 3) We showed, through a manual review of radiographs, that the accuracy for the CNN differed from the computed base accuracy and was considerably higher. We showed that LDA was not well suited for labeling images based on radiologist reports.

*Topics and Text Analysis of Radiologist Reports*

LDA was designed for a corpus containing individual documents assumed to be about a central topic and subtopics, i.e., of some length [26,41]. Radiologist reports are concise with a restricted specialist vocabulary and relatively standardized language but variable in scope and length due to their dependency on the exam referral. Experiment 1 showed that the best way to model our data was to use individual sentences as documents rather than reports, with fewer expected topics per document.

ML algorithms thrive on data, and more data is usually better. The theory for LDA and topic modeling generally holds that longer documents and more data are better as they depend on comparing how words co-occur between documents [41,42]. Short documents decrease the likelihood of words co-occurring between documents (i.e., giving sparse data). Yan et al. [41] and Quercia et al. [42] gave examples of topic models for short documents, but these have not been widely adopted or extensively tested. Even so, for our corpus, we found that less text was better for finding concrete and usable topics, and we could adapt our LDA topic model to find topics within our corpus. It was likely due to the homogeneity, focus, and limited specialist vocabulary in radiologist reports, meaning that the word co-occurrence was significant even for short documents. It motivates future studies, and the results could be useful when modeling other journal data. Dividing the whole dataset into separate anatomies (wrist and ankle) and modeling them separately was consistent with the divide-and-conquer results from experiment 1. It was also shown to be the best approach in a pilot study. There is a risk that specific and common topics for one anatomy might get lost for the model as a whole. A hypothetical topic describing a fracture in the talus (a bone connecting the ankle and foot) found in 20% of the ankle radiographs would make up 5% of the corpus if wrists had twice the number of images. If we always guessed that there was no fracture of the talus, we would have an accuracy of 80% in the first case (only ankle radiographs) and 95% accuracy in the second case (ankle and wrist radiographs). This motivated the use of less data to construct useful topics. This held for anything but broad topics, like fractures in general.

There is a trade-off between the ML preference for more data and many topics where many would be closely related [28], making it difficult to interpret our approach. In experiment 2, we examined individual anatomies, which gave fewer and more distinct individual topics but a smaller corpus. Our more incremental approach, where each anatomy is included separately, could allow for incremental improvements and additions of anatomies. In time, a considerable amount of corpus could be generated and reused in later implementations. On the other hand, this does not leverage the full power of ML. For example, based on clinical experience, arthrosis is not commonly asked for in hand examinations but more frequently in hip examinations. The ML algorithm will have less information from which to detect arthrosis in hand data than hip data using our segmental approach, meaning that arthrosis of the hand could become undetectable. In an approach combining all anatomies, arthrosis in the hand would likely be detected along with arthrosis in the hip, as it looks and is described similarly across the entire body.

*Loss of Semantic Context*

The bag-of-words model means that crucial syntactic information is lost. An example would be a topic that can contain its negation, e.g., "no fracture, but visible luxation" and "fracture but no visible luxation." For this kind of example, LDA could detect and separate the topics of fracture and luxation but was poor at separating positivity or negativity for respective topics. A nuance would be a document stating "no additional fracture present," indicating that some fracture is present but could be interpreted as no fracture. LDA did not capture these nuances well, but by studying sentences rather than whole reports, we hoped to overcome them when we aggregated the results from individual sentences in a report. This could have been dealt with using sentiment analysis, an NLP technique determining whether a text is positive or negative about its topic [43]. Combining sentiment analysis with our topic labels could be potent but requires thousands of manually annotated reports.

*CNN Accuracy as a Proxy for LDA Utility*

This study did not examine the LDA topics except for generating labels, i.e., a tool for automated feature extraction from texts to train a CNN. This was in keeping with Shin et al. and other studies in the field using LDA for image labeling [29,44,45]. While similar, the stochastic nature of LDA means that if the same experiments were to be rerun, they would not give the same distributions [26,28]. A rigorous study of the LDA outcomes would lack generality. While label quality is essential, our results show perfect labels are not.

*CNN Performance for Radiographs*

Our study partially paralleled the procedure from Shin et al. [29]. We focused on the clinical outcome regarding pathology detection in radiographs and CNN accuracy rather than the LDA outcomes and, therefore, introduced a manual review process to estimate the method's true performance. We achieved considerably better results than Shin et al. for CNN on some labels, but we also had more images in a more limited domain. For experiment 3, we manually reviewed the radiographs, compared them to a gold standard, and found that the CNN performed even better than the base $accuracy_{base}$ indicated. For example, for fractures, we showed that the true accuracy was 16% higher than the base accuracy indicated and 8% higher for fibular fractures. This shows a considerable number of labeling errors and illustrates the limits of LDA in analyzing radiographic reports. Our CNN could learn and discriminate against these image features despite this. In an unpublished study, currently in the manuscript, we achieved an accuracy of 86% (95%CI 83.5% to 90.0%) for fracture detection in radiographs, where text searches in reports had been used to create the image labels. Our study reached 91% (95%CI 87.8% to 93.7%) accuracy. While the LDA-derived labels could be used to train the CNN to detect features in the images, our results were not significantly better than those of labeling reports using text searches and more straightforward methods. For other topics, the LDA-trained CNN did not exceed the mode, and constantly guessing the mode (usually false) would have given a better accuracy. For styloid fractures, the accuracy was dismal. In a study by Wang et al. [46], using a different but related CNN approach to detect malignant features in mammographic images, they reached accuracies of 61.3% and 89.7% for detecting tumor mass and calcifications, respectively, in 201 test images, making the outcomes comparable to ours.

Considering our results and the study by Shin et al. [29], we conclude that while it was possible to use, it was not superior to other methods (such as text matching) for labeling radiographic images from reports. However, considering the study by Wang [46], we found that our results were on par with CNN image analysis of radiographs.

*Analysis of Other Topics*

The OD proved unexpectedly difficult. With an OD, it is seemingly evident that there has been a fracture at some point, but it was not always visible as the foreign material sometimes obscured the fracture on the radiograph. Such devices were counted as fractures by the network and the reviewer, even though this was not necessarily true as they could be remnants from a previous fracture or an arthrodesis. This was also confounding in some cases as radiographs are two-dimensional see-through projections of a three-dimensional space, where everything is superimposed on top of everything else. For example, in a side view of a fibula fracture, the OD would seem to be fixed to the fibula. Still, from a different view (a frontal view), the device was part of the tibia and was only superimposed onto the fibula. The same problem was present with casts, as these were almost exclusively labeled as having whatever fracture class was reviewed, even when the fracture was obviously elsewhere. It was rare that a casting was labeled as not having a fracture visible. Casts were also present in many images incorrectly classified as having an OD, with a high number of false positives. Casts will likely occur along with OD postoperatively and can look like external fixations. They were, however, not always classified as such, and there were a few false negatives where casting and OD occurred together.

Not visible in the numbers of Table 7 is that the accuracy for fibular fracture would be lower if not for the fact that these fractures tended to coincide with other fractures. It was clear from the manual review that the CNN reacted positively to any indication of a fracture anywhere in the image, and $accuracy_{true}$ which would have been considerably lower if not for this concurrence. Given that the fibular fracture category was similar in size to the OD category and reached much better results, this hid poor performance.

Compared to the gold standard, there were fewer errors among the correctly classified images in general, but, as expected, classification errors were more common among the misclassified images. We perceived that misclassified images were, in general, more difficult to review, taking more time and requiring more scrutiny. One explanation could be that these are difficult to classify and that the report contains uncertainty captured by the LDA but not the image review. For example, "probable fracture," "likely fracture," or "unlikely fracture" all indicate ambiguity and could indicate anamnestic information from the referral or just that it is challenging to tell from the radiograph. However, we did not show this conclusively.

## Strengths and Limitations

This study utilized a unique dataset of 88,000 reports and 235,000 radiographic images. Extending the methods to more anatomies and millions of exams is straightforward. We also compared selected labels to gold standards of 300 images per label, for a total of 1,500 images, giving a reasonable estimate of the CNN's accuracy at detecting each label in the radiographs.

To our knowledge, our study was the first to explicitly study LDA and CNN for orthopedic trauma radiography and attempt to optimize it for radiologist reports in the Swedish language. In the process, we showed the need for calibration. We also generated a gold standard for our primary outcome (CNN outcome for the five selected labels) and manually studied many radiographs.

We customized a freely available CNN and trained it to detect image features in radiographs. This shows that these powerful methods could be available to the average clinician with limited training at minimal cost.

LDA was founded on the idea of documents with a main topic and several minor topics, i.e., a longer text [26]. The radiologist reports in this study did not conform to this ideal. "No fracture" is a valid and highly relevant complete two-word report that was not uncommon in our data. While there are LDA models that try to deal with short documents [42,47], we did not implement such models as they were special-purpose implementations unsuited for our task or not sufficiently studied. Instead, we tried to accommodate this by calibrating the LDA. The bag-of-word model generated by LDA is good at finding topics [26,27,48] but not optimal for data labeling.

Radiologist reports are answers to exam referrals containing questions (usually diagnoses) posed by physicians. Some information in the image is omitted from the report or has no bearing on the referral. This information, for example, a pathology, is not mentioned and is therefore impossible to extract for image labeling, a one-sided misclassification error. This will be more common for rare outcomes and those that were common but rarely asked for in trauma referrals, such as osteoarthritis. We tried to compensate for this by manually reviewing positive and negative outcomes.

A radiographic examination entails several images in different projections. Our CNN looked at one image at a time. The image feature deduced from the report would be valid for the examination but not necessarily for every radiograph. We compensated for this by creating a gold standard for radiographs to determine the accuracy of the CNN. From the manual review, we gleaned that this was indeed an issue but could not assess to what extent.

While using the referral and patient records for each examination would be incredibly valuable, they contain personal information that would be difficult to anonymize. It was not within the remit of our ethical approval.

## Clinical Significance

We automatically extracted clinically relevant information from medical reports using ML and transferred it into automated image interpretations by a CNN. We showed how considering the clinical nature of the reports was essential to applying LDA to topic modeling. Despite calibration, LDA was insufficient for labeling images, giving a high degree of labeling errors; however, we showed that the CNN trained from LDA labels still had a high accuracy. The implication is that we have a computerized system to automatically find noteworthy themes in text and then correlate that information to features in images, i.e., make its interpretation of medical data – the foundations of artificial intelligence.

## Future Use and Studies

Finding ways to improve the performance of the CNN, for example, by considering all images in an examination, could improve the method's accuracy. Future research could include auto-generated image captions (as in Vinyals [39].) These draft radiologist reports could speed up the work of radiologists by providing a prediction of what they will write. They could also help label individual images instead of entire examinations with multiple images in different projections, which could be used for improved LDA labeling. They could also help amend the problem of missing information in the reports, i.e., the one-sided classification error. With improved accuracy, the results obtained by the CNN could be used in the clinic as a fast-screening tool. CNN image analysis and retrospective studies could be used for clinical review. Ultimately, finding ways to detect and predict how a pathology will evolve, for example, if a fracture will dislocate or not, and suggest evidence-based treatment would benefit patients, doctors, and society. It could also be used to find new predictive features from the radiographs. Beyond that, extending ML and AI, as studied here, to other fields of medicine, such as mammographic imaging, could provide new insights into medicine.

## Conclusions

CNNs were suited to study orthopedic trauma radiographs. In addition, we found that we could extract clinically relevant features from the radiologist reports using LDA and automatically categorize thousands of images according to these labels. We could successfully train the CNN to detect the, by LDA, autodetected image features despite the high level of labeling errors primarily due to exam-level reports for single images and lack of referral information. While not superior to other methods, a standard implementation of LDA could be adapted to become a helpful tool for labeling orthopedic radiographs from reports. The study also illustrates how ML and CNNs could be transferred to medical research. It could enable new and potentially revolutionary research if used by ordinary clinicians.

## Acknowledgments, Ethics, and Disclosures

The Swedish Association of Local Authorities and Regions funded this project and the institution where it was performed. We gratefully acknowledge the support of NVIDIA Corporation, which donated two Tesla K40 GPUs to this research. We are grateful for Nicklas Fahlberg's help in labeling topic data.

The Stockholm Ethical Review Board has deemed the research not subject to special ethical review (Dnr. 2014/453-31/3) because, after anonymization, the images and reports are not considered human subjects.

The authors had no conflicts of interest. This article is an abridged version of JO's 2016 master's thesis, available from the Karolinska Institute or the corresponding author upon request.

*CRediT (https://credit.niso.org/)[49]*

JO: Data curation, formal analysis, investigation, methodology, project administration, software, validation, visualization, writing - original draft, writing – review & editing
MG: Conceptualization, funding acquisition, methodology, resources, software, supervision, validation, writing – review & editing

# Supplement

**Table S1: Example of LDA-generated topics** generated in experiment 2. Top words (with more than 3% incidence for the topic) and top 15 sentences were displayed. These were then translated into labels with features that could be found in a radiograph.

| Top words (%) | Top 15 sentences per topic | Labels |
|---|---|---|
| Top 89.7%: radius (27.6%), distalt (14.7%), fraktur (13.6%), ulna (10.7%), distal (10.3%), tvärfraktur (7.6%) | Tvärfraktur distala ulna med liten dislokation<br>Distal underarmsfraktur med snett förlopp helt distalt i radius och ulna<br>Distal underarmsfraktur med snett förlöpande frakturer helt distalt i ulna och radius<br>Tvärfrakturer distalt radius och ulna<br>Vertikalt löpande frakturspalt distalt i radius utan dislokation<br>Tvärfrakturer i distala radius och ulna<br>Färsk fraktur distalt i radius med 25 grader dorsalbockning relativt radius längsaxel<br>Distal underarmsfraktur endast någon centimeter från ledytorna distalt i radius och ulna<br>Gamla läkta distala frakturer i radius och ulna<br>Tvärfraktur distalt radius med isärsprängt distalt fragment<br>Tvärfrakturer distalt i radius samt ulna<br>Fraktur i caput radii med 2 millimeters nedpressning utav ledytan radiellt<br>Fraktur i distala radius med minimal dislokation<br>Man har nu också applicerat 2 tvärgående stift distalt över radius och ulna<br>I prelimärsvaret anges fraktur i distala radius med minimal dislokation men fyndet får anses vara mycket osäkert | fraktur |
| Top 97.5%: distal (37.5%), radiusfraktur (32.9%), intraartikulär (17.2%), komminut (10.0%) | Komminut distal intraartikulär distal radiusfraktur med kraftig dorsalvinkling och dorsalförskjutning av distala fragmenten tillsammans med carpus<br>Distal intraartikulär komminut radiusfraktur med distala fragmentet ventralförskjutet en dryg kortikalisbredd<br>Komminut intraartikulär distal radiusfraktur med kraftig felställning<br>Komminut distal radiusfraktur med intraartikulära komponenter<br>Komminut intraartikulär distal radiusfraktur med bland annat uppsplittrat bakstöd<br>Distal komminut intraartikulär radiusfraktur<br>Komminut intraartikulär distal radiusfraktur<br>Distal komminut intraartikulär radiusfraktur med uppsplittrat bakstöd<br>Comminut intraartikulär distal radiusfraktur med volar- och radialkompression<br>Distal intraartikulär komminut radiusfraktur<br>Distal radiusfraktur, komminut med intraartikulär komponent<br>Distal intraartikulär något komminut radiusfraktur<br>Distal, intraartikulär, komminut radiusfraktur med vertikal utlöpare<br>Komminut distal intraartikulär radiusfraktur<br>Komminut distal radiusfraktur med en intraartikulär utlöpare | Fraktur<br><br>radius fraktur<br><br>distal radius fraktur<br><br>komminut fraktur |